\documentclass[10pt,twocolumn,letterpaper]{article}

\usepackage[pagenumbers]{cvpr} 

\usepackage{graphicx}
\usepackage{times}
\usepackage{helvet}
\usepackage{courier}
\usepackage{amsmath}
\usepackage{algorithm}
\usepackage{algorithmic}
\usepackage{csquotes} 
\usepackage{paralist}
\usepackage{amssymb}
\usepackage{indentfirst}
\usepackage{float}
\usepackage{multirow}
\usepackage{cite}

\usepackage{color}
\usepackage{xcolor}

\usepackage{mathrsfs}
\usepackage{textcomp,booktabs}
\usepackage{amssymb}
\usepackage{pifont}
\usepackage[misc]{ifsym}
\newcommand{\cmark}{\ding{51}}%
\newcommand{\xmark}{\ding{55}}%

\usepackage{mathrsfs}

\definecolor{citecolor}{HTML}{0071bc} 
\definecolor{SeaGreen4}{RGB}{0,205,102} 
\definecolor{SlateBlue}{RGB}{106,90,205} 
\definecolor{DarkRed}{RGB}{178,34,34} 
\usepackage[colorlinks, linkcolor=red,  anchorcolor=blue, citecolor=citecolor]{hyperref}

\usepackage{colortbl}
\definecolor{mygray}{gray}{.9}
\definecolor{mypink}{rgb}{.99,.91,.95}
\definecolor{mycyan}{cmyk}{.3,0,0,0}

\definecolor{citecolor}{HTML}{0071bc} 
\definecolor{SeaGreen4}{RGB}{0,205,102} 
\definecolor{SlateBlue}{RGB}{106,90,205} 
\definecolor{DarkRed}{RGB}{178,34,34}

\usepackage[capitalize]{cleveref}
\crefname{section}{Sec.}{Secs.}
\Crefname{section}{Section}{Sections}
\Crefname{table}{Table}{Tables}
\crefname{table}{Tab.}{Tabs.}

\definecolor{cvprblue}{rgb}{0.21,0.49,0.74}

\def\paperID{9748}      
\def\confName{CVPR}
\def\confYear{2024}

\title{ Event Stream based Human Action Recognition: A High-Definition Benchmark Dataset and Algorithms }

\author{Xiao Wang$^{1}$, Shiao Wang$^{1}$, Pengpeng Shao$^{2}$, 
    Lin Zhu$^{3}$, Bo Jiang$^{1}$\thanks{Corresponding Author: Bo Jiang}, Yonghong Tian$^{4,5,6}$ \\ 
${^1}${School of Computer Science and Technology, Anhui University, Hefei, China} \\
${^2}${Tsinghua University, Beijing, China}, ~~  
${^3}${Beijing Institute of Technology, Beijing, China} \\
${^4}${Peng Cheng Laboratory, Shenzhen, China}, ${^5}${School of Computer Science, Peking University, China} \\ 
${^6}${Shenzhen Graduate School, Peking University, China} \\ 
\textit{xiaowang@ahu.edu.cn}, \textit{wsa1943230570@126.com}, \textit{zeyiabc@163.com}, \\ 
\textit{ppshao@tsinghua.edu.cn}, \textit{\{linzhu, yhtian\}@pku.edu.cn} \\ 
\url{https://github.com/Event-AHU/CeleX-HAR}
}

\begin{document}
\maketitle

\begin{abstract}
Human Action Recognition (HAR) stands as a pivotal research domain in both computer vision and artificial intelligence, with RGB cameras dominating as the preferred tool for investigation and innovation in this field. However, in real-world applications, RGB cameras encounter numerous challenges, including light conditions, fast motion, and privacy concerns. Consequently, bio-inspired event cameras have garnered increasing attention due to their advantages of low energy consumption, high dynamic range, etc. Nevertheless, most existing event-based HAR datasets are low resolution ($346 \times 260$). In this paper, we propose a large-scale, high-definition ($1280 \times 800$) human action recognition dataset based on the CeleX-V event camera, termed CeleX-HAR. It encompasses 150 commonly occurring action categories, comprising a total of 124,625 video sequences. Various factors such as multi-view, illumination, action speed, and occlusion are considered when recording these data. To build a more comprehensive benchmark dataset, we report over 20 mainstream HAR models for future works to compare. In addition, we also propose a novel Mamba vision backbone network for event stream based HAR, termed EVMamba, which equips the spatial plane multi-directional scanning and novel voxel temporal scanning mechanism. By encoding and mining the spatio-temporal information of event streams, our EVMamba has achieved favorable results across multiple datasets. Both the dataset and source code will be released. 
\end{abstract}

\section{Introduction}

Human Activity Recognition (HAR) represents a critical domain within computer vision that has experienced substantial advancement in recent times, thanks to the integration of deep learning techniques~\cite{ahmad2021graph, kong2018humanARSurvey}. Typically, these models are tailored for analyzing video frames recorded by RGB cameras and have been extensively employed across a broad spectrum of practical scenarios. For example, we can achieve pre-prevention, in-process monitoring, and post-inspection in the security monitoring field, and intelligent referee in sports through the analysis of human behavior. Although the RGB cameras based HAR works well in regular scenarios, however, the issues caused by its imaging quality may limit the applications of HAR severely, such as low illumination and fast motion. On the other hand, the privacy protection is also widely discussed in the human-centered research. Awkwardly, \emph{the ethical problems} caused by high-quality data and \emph{the data quality problems} caused by low-quality video both require new behavior recognition paradigms.

\begin{figure*}[!htb]
\center
\includegraphics[width=7in]{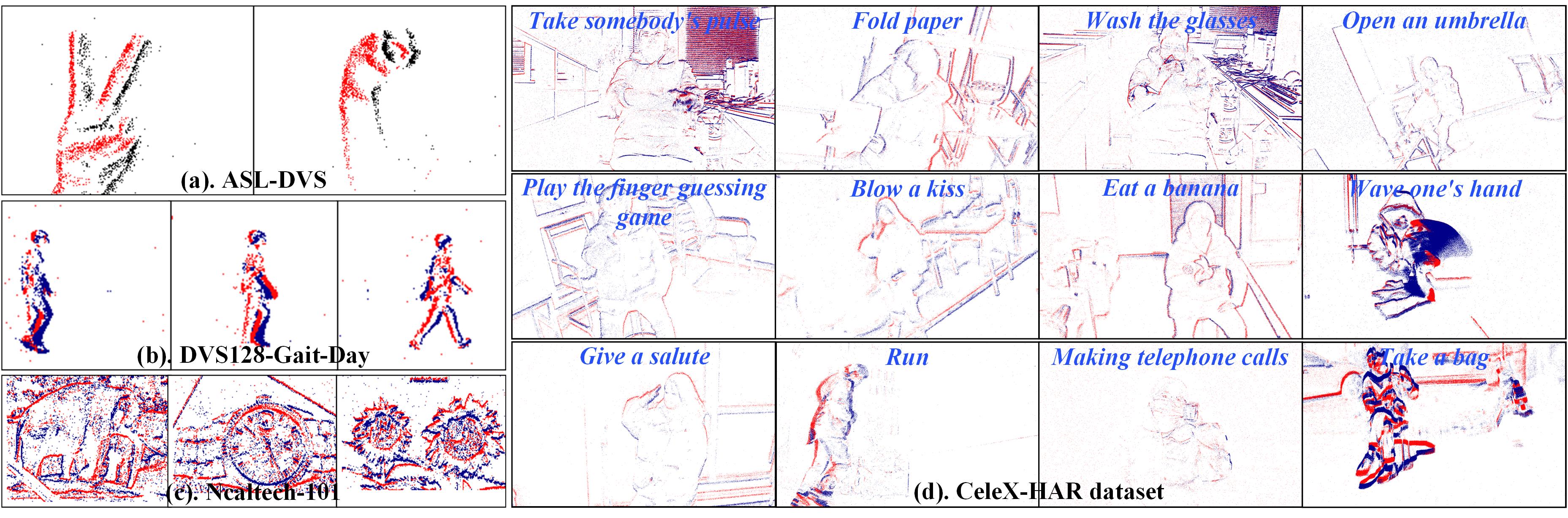} 
\caption{Comparison of existing Event datasets (a). ASL-DVS, (b). DVS128-Gait-Day, (c). N-Caltch101, and (d) our newly proposed CeleX-HAR dataset.}  
\label{fig::firstIMG}
\end{figure*}

Recently, the event camera (also termed Dynamic Vision Sensors, DVS) which is a bio-inspired sensor draws more and more attention from researchers. Different from the RGB camera which records the scene into video frames in a synchronous way, each pixel in the event camera is triggered asynchronously by saving an event point if and only if the variation of intensity exceeds the given threshold. Due to the aforementioned unique imaging principle, the event camera shows the following advantages or features: \emph{high dynamic range}, \emph{low energy-consumption}, \emph{dense temporal resolution but sparse spatial resolution} \cite{gallegoevent}. Therefore, it performs well even in low-illumination, overexposure, and fast-motion scenarios. Also, the spatial resolution is getting higher, for example, $1280 \times 800$ and $1280 \times 720$ can be achieved by the CeleX-V~\cite{chen2019CeleXV} and PROPHESEE, respectively. These features all inspired us to address the pain points of HAR using a high-resolution event camera.

Indeed, several datasets are proposed for the event-based classification~\cite{orchard2015converting, serrano2015poker, li2017cifar10, amir2017low, miao2019neuromorphic, lin2021esimagenet, kim2021n} before,  but they are either simulation data or obtained by recording the screen, thus it's intractable to reflect the true features of event camera fully. For the realistic ones~\cite{LiuXTM021ijcai, bi2020graph}, their resolution, length or scale is limited (ASL-DVS~\cite{bi2020graph}: $240 \times 180$,  DailyAction~\cite{LiuXTM021ijcai}: $346 \times 260$), and the overall performance is almost saturated (TORE~\cite{baldwin2022TORE} achieves 0.9995 on ASL-DVS~\cite{bi2020graph}, 0.945 on N-Cars~\cite{sironi2018hats}). Existing works~\cite{liu2024VMamba} and our experimental results demonstrate that current network architectures perform well on low-resolution data (e.g., $224 \times 224$), but they achieve inferior results on high-resolution ones. 
This raises an interesting and contemplative research problem. We hypothesize that early in the deep learning boom, given the computational constraints, successful models were optimized for low-resolution data. As computing resources and sensor technology have evolved, high-definition data has become increasingly widespread. Continuing to rely on low-resolution data, and disregarding the valuable information in high-definition data, would be a misuse of resources. 
Therefore, a high-quality, high-resolution event-based action recognition dataset is urgently needed in the current academic community, as it can effectively support and promote the design of network structures for high-resolution input signals.

\begin{table*}
\center
\caption{\textbf{Comparison of Event datasets for Human Action Recognition.} M-VW, M-ILL, M-MO, DYB, OCC, and DR denote multi-view, multi-illumination, multi-motion, dynamic background, occlusion, and duration of the action, respectively. Note that we only report these attributes of realistic DVS datasets for HAR.} \label{datasetlist} 
\resizebox{\textwidth}{!}{ 
\begin{tabular}{l|ccccccccccccc}
\hline \toprule [0.5 pt]
\textbf{Dataset}    &\textbf{Year}  &\textbf{Sensors}   &\textbf{Scale}   &\textbf{Class}  &\textbf{Resolution}  &\textbf{Real}  &\textbf{M-VW} &\textbf{M-ILL}  	&\textbf{M-MO}  &\textbf{DYB} &\textbf{OCC} &\textbf{DR} \\ 
\hline 
\textbf{ASLAN-DVS}~\cite{lin1811temporal}   	 	&2011 	&DAVIS240c    &$3,697$ 		&432 			&$240 \times 180$ 		&\xmark	&-    &- 		&- 		&-		&- 	&-  	\\ 
\textbf{MNIST-DVS}~\cite{serrano2015poker}   		 &2013   				&DAVIS128    &$30,000$ 		&10 		&$128\times128$ 			&\xmark  			&-    &- 		&- 		&-		&- 	&-  	\\ 
\textbf{N-Caltech101}~\cite{orchard2015converting}  &2015  &ATIS    &$8,709 $		&101 		&$302 \times 245$ 			&\xmark 	 &-    &- 		&- 		&-		&- 	&-  \\ 
\textbf{N-MNIST}~\cite{orchard2015converting}   &2015  	&ATIS     &$70,000$ 		&10 		&$28\times28$ 			&\xmark  			&-    &- 		&- 		&-		&- 	&- 	\\ 
\textbf{CIFAR10-DVS}~\cite{li2017cifar10}   		 &2017 		&DAVIS128   &$10,000$ 		&10 		& $128\times128$			&\xmark  			&-    &- 		&- 		&-		&- 	&-  	\\ 
\textbf{HMDB-DVS}~\cite{kuehne2011hmdb}   	&2019  	&DAVIS240c    &$6,766$ 		&51 		&$240 \times 180$ 			&\xmark    &-    &- 		&- 		&-		&- 	&- 	\\ 
\textbf{UCF-DVS}~\cite{soomro2012dataset}   	&2019  	&DAVIS240c    &$13,320$  		&101 		&$240 \times 180$ 			&\xmark    &-    &- 		&- 		&-		&- 	&-  	\\ 
\textbf{N-ImageNet}~\cite{kim2021n} &2021 	&Samsung-Gen3    &$1,781,167$   &1000   &$480 \times 640$   &\xmark  &-    &- 		&- 		&-		&- 	&- 	\\  
\textbf{ES-ImageNet}~\cite{lin2021imagenet}  &2021 	&-    &$1,306,916$  &1000  		&$224 \times 224$  	 &\xmark  &-    &- 		&- 		&-		&- 	&- 	\\ 
\textbf{N-EPIC-Kitchens}~\cite{plizzari2022e2} &2022 	&-    &$10,000$   &-   &-   &\xmark  &-    &- 		&- 		&-		&- 	&- 	\\  
\textbf{N-ROD}~\cite{cannici2021nrod} &2022 	&-    &41,877   &51   &$640 \times 480$   &\xmark  &-    &- 		&- 		&-		&- 	&- 	\\ 
\hline 
\textbf{DvsGesture}~\cite{amir2017low}   	&2017 		&DAVIS128   &$1,342$ 		&11 		&$128 \times 128$ 		&\cmark  			&\xmark    &\cmark 		&\xmark 		&\xmark			&\xmark  	& - 		\\ 
\textbf{N-CARS}~\cite{sironi2018hats} &2018  	&ATIS    &$24,029$ 		&2 		&$304 \times 240$  		&\cmark 	&\cmark    &\xmark 		&\cmark 		&\xmark			&\cmark 		 	&-  		\\ 
\textbf{ASL-DVS}~\cite{bi2020graph}	&2019 	&DAVIS240    &$100,800$ 		&24 			&$240 \times 180$ 		&\cmark  	&\xmark    &\xmark  &\xmark 		&\xmark 	&\xmark  	&0.1s  	\\ 
\textbf{PAF}~\cite{miao2019neuromorphic}	&2019 	&DAVIS346    &$450$ 		&10 			&$346 \times 260$ 		&\cmark  	&\xmark    &\xmark  &\xmark 		&\xmark 	&\xmark  	&5s  	\\ 
\textbf{DailyAction}~\cite{liu2021event}  &2021 	&DAVIS346    &$1,440$ 		&12 			&$346 \times 260$ 		&\cmark  	&\cmark    &\cmark  &\xmark 		&\xmark 	&\xmark  	&5s  	\\ 
\textbf{Bully10K}~\cite{dong2024bullying10k} &2023 		&DAVIS346    &$10,000$  &10		&$346 \times 260$ 		&\cmark  &\cmark     &\cmark 		&\cmark 		&\xmark			&\xmark	  	& 2-20s  	\\ 
\textbf{HARDVS}~\cite{wang2024hardvs} &2024 		&DAVIS346    &$107,646$ 		&\textbf{300}		&$346 \times 260$ 		&\cmark  &\cmark    &\cmark 		&\cmark 		&\cmark			&\cmark 	  	&5s  	\\
\textbf{PokerEvent}~\cite{wang2023sstformer} &2024 		&DAVIS346    &$27,102$ 		&114		&$346 \times 260$ 		&\cmark  &\cmark    &\cmark &\cmark &\cmark	&\cmark 	  	&-  	\\
\textbf{THU$^{E-ACT}$-50}~\cite{gao2023action} &2024  &CeleX-V &$10,500$  &50	&$1280 \times 800$  	&\cmark   &\cmark    &\cmark 		&\cmark 		&\xmark			&\xmark 	  	&-  	\\ 
\textbf{Dailydvs-200}~\cite{wang2024dailydvs} &2024  &DVXplorer Lite &$22,046$  &200	&$320 \times 240$  	&\cmark   &\cmark    &\cmark 		&\cmark 		&\cmark		&\cmark 	  	&1-20s  	\\ 
\hline 
\textbf{CeleX-HAR} \textbf{(Ours)}  &2024 		&CeleX-V    &$\textbf{124,625}$  &150		&$\textbf{1280} \times \textbf{800}$ 		&\cmark  &\cmark    &\cmark 		&\cmark 		&\cmark			&\cmark 	  	&2-3s   	\\ 
\hline \toprule [0.5 pt]
\end{tabular}
} 
\end{table*}

To bridge the data gap, we propose a large-scale benchmark dataset for event-based HAR in this work, termed CeleX-HAR. As shown in Fig.~\ref{fig::firstIMG}, CeleX-HAR contains 124,625 high-resolution event streams ($1280 \times 800$), collected using a CeleX-V event camera and covers 150 categories of human daily activities, such as \emph{moving a chair, putting on shoes, opening an umbrella}, etc. Furthermore, CeleX-HAR considers different attributes, such as multi-view, various illumination conditions, camera motions, speed of action, occlusion, glitter, and capture distance. We split the CeleX-HAR dataset into a training and testing subset which contains 99,642 and 24,983 videos, respectively. Due to the scarcity of baseline methods for future comparison, we have trained and reported on over 20 recognition models on our dataset to facilitate better the comparison and development of event data-based action recognition models. We anticipate that our newly introduced high-definition CeleX-HAR dataset will stimulate and foster research in this field.

Based on the newly proposed CeleX-HAR dataset, we further propose an event spatial-temporal scanning mechanism-guided Mamba framework for human action recognition, termed EVMamba. The key insight of this framework is that existing action recognition algorithms primarily focus on learning local features only using convolutional neural network (CNN)~\cite{he2016deep}, but ignore the long-range relations. Visual Transformers~\cite{dosovitskiy2020ViT} excels at capturing long-range dependencies, however, they are computationally expensive ($\mathcal{O}(N^2)$) and memory-intensive,  which presents significant challenges for practical deployment. The newly introduced Mamba model, which maintains a good balance between precision and computational cost with a complexity of $\mathcal{O}(N)$, and has been adapted for video classification tasks~\cite{li2024videomamba, lu2024videomambapro}. Nonetheless, there has been no research on adapting it for human action recognition using event cameras. The output event streams, similar to 3D point clouds, necessitate efficient tokenization. This is a crucial factor that allows sequential models to function optimally in the field of HAR. Specifically, as shown in Fig.~\ref{fig:framework}, given the event streams, we propose to extract the tokens from both spatial and temporal views using the newly proposed event spatial-temporal scanning mechanism. It conducts a spatial cross-scan to get the event frame tokens and a temporal voxel scanning for temporal token mining. These tokens are transformed using patch and voxel embedding layers and fed into the VMamba blocks for spatial-temporal feature learning. Finally, we add these two features and feed them into the classification head for action recognition.

To sum up, the contributions of this paper can be summarized as the following three aspects: 

$\bullet$ We propose a large-scale benchmark dataset for event-based human activity recognition, termed CeleX-HAR. To the best of our knowledge, it is the largest high-resolution event dataset for human action recognition.  

$\bullet$ We further present an event recognition framework based on the Visual Mamba architecture and develop an innovative voxel temporal scanning mechanism, it strikes a good balance between the model complexity and recognition performance. 

$\bullet$ We train and report more than 20 classification models on the proposed CeleX-HAR dataset, which provides a good platform for subsequent works to compare. Extensive experiments conducted on CeleX-HAR and multiple other widely used benchmark datasets fully demonstrate the effectiveness of the proposed model.

\section{Related Work}
In this section, we review the most related research topics to our paper, including Event-based Recognition and the State Space Model. More related works can be found in surveys~\cite{kong2022human, sun2022human, zhang2019comprehensive, zhu2016handcrafted}.

\subsection{Event-based Recognition} 
Current works can be divided into three streams for the event-based recognition, including the CNN based~\cite{wang2019evGait}, spiking neural network (SNN) based~\cite{fang2021PLIF, fang2021SNNIIR}, graph neural network (GNN) based models~\cite{bi2019gnnevent, bi2020graph}, due to the flexible representation of event stream. 
For the CNN based models, Wang et al.~\cite{wang2019evGait} propose to identify human gaits using an event camera and design a CNN model for recognition. 
As the third generation of neural networks, the SNN is also adopted to encode the event stream for energy-efficient recognition. To be specific, 
Peter et al.~\cite{diehl2015snnbalancing} propose the weight and threshold balancing method to achieve efficient ANN-to-SNN conversion. 
Nicolas et al.~\cite{perez2021sparse} propose a sparse backpropagation method for SNNs and achieve faster and more memory efficiency. 
Zhou et al.~\cite{zhou2024exact} propose ExACT, a novel approach to event based action recognition through cross-modal conceptualization, along with an adaptive fine-grained event representation and uncertainty estimation module based on conceptual reasoning.
Gao et al.~\cite{gao2023action} proposes an event camera based behavior recognition framework EV-ACT, which integrates multiple event information through Learning Multi Fused Representation (LMFR). Gao et al.\cite{gao2024hypergraph} propose a multi-view event camera action recognition framework HyperMV based on hypergraphs, which captures the relationship between cross-view and temporal features by constructing a multi-view hypergraph neural network.

For the point cloud based representation, Wang et al.~\cite{wang2019spaceCloud} treat the event stream as space-time event clouds and adopt PointNet~\cite{qi2017pointnet} as their backbone for gesture recognition. 
Sai et al.~\cite{vemprala2021representation} propose the event variational auto-encoder (eVAE) to achieve compact representation learning from the asynchronous event points directly. 
Fang et al.~\cite{fang2021snnresnet} propose SEW (spike-element-wise) residual learning for deep SNNs which addresses the vanishing/exploding gradient problems effectively. 
Meng et al.~\cite{meng2022DSR} propose an accurate and low latency SNN based on the Differentiation on Spike Representation (DSR) method. 
TORE~\cite{baldwin2022TORE} is short for Time-Ordered Recent Event (TORE) volumes, which compactly stores raw spike timing information. 
VMV-GCN~\cite{xie2022vmv} is proposed by Xie et al. which is a voxel-wise graph learning model to fuse multi-view volumetric. 
Li et al.~\cite{li2022eventFormer} introduce the Transformer network to learn event-based representation in a native vectorized tensor way. 
Different from these works, in this paper, we design a novel event spatial-temporal scanning mechanism to adapt the VMamba network for the action recognition task effectively and efficiently.

\begin{figure*}
\centering
\includegraphics[width=1\linewidth]{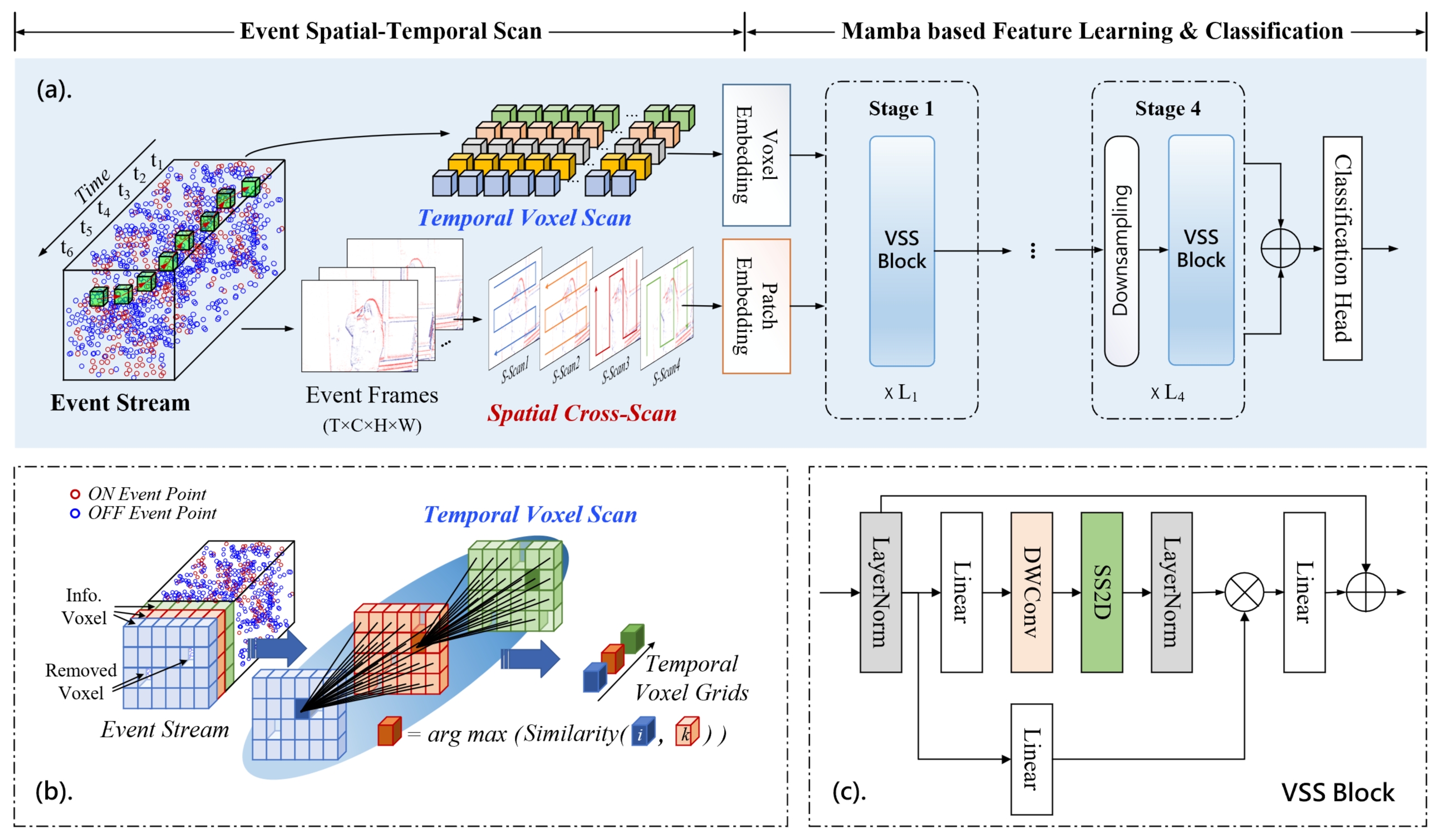}
\caption{An overview of our proposed EVMamba for event-based human action recognition. Given the event streams, we design a novel spatial-temporal scanning mechanism that mines both spatial tokens and temporal voxel tokens as the input of the Mamba network. Enhanced by the temporal voxel scanning strategy, our EVMamba outperforms the baseline on multiple event-based classification benchmark datasets.} 
\label{fig:framework}
\end{figure*}

\subsection{Human Action Recognition}  
Human action recognition(HAR) has been a popular task in the past decade due to its practicality in the real world. It is a fundamental task that recognizes a human action from a video containing complete action execution. In deep learning, CNN is widely used in computer vision, including action recognition. Wang et al.~\cite{wang2016action} proposes an effective method to encode the spatiotemporal information into color distributions in three two-dimensional images, learning discriminative features for human action recognition through CNN. Soo et al.~\cite{soo2017interpretable} propose a new model for 3D human motion recognition based on a Time Convolutional Neural Network (TCN). This model can interpret 3D bones and easily understand spatiotemporal representations. Li et al.~\cite{li2017joint} utilize CNN to process the spatiotemporal information of skeleton sequences, significantly improving the effectiveness of human action and interaction recognition in single and cross-view scenarios. Sudhakaran et al.\cite{sudhakaran2023gate} propose a novel spatiotemporal feature extraction module called Gate Shift Use (GSF) to improve the performance of 3D CNN in video action recognition. 

With the emergence of Transformer networks in the past few years, many Transformer based HAR methods have emerged. Wang et al.~\cite{wang2024hardvs} introduce a new framework for human activity recognition based on dynamic visual sensors. The event stream is mapped to spatiotemporal embedding through StemNet, and a Transformer network is used to encode and decode dual-view representations.  Li et al.~\cite{li2023semantic} propose a new pattern recognition framework that integrates RGB/Event features and semantic features using a multimodal Transformer network, thereby solving the semantic gap and small-scale backbone network problems in existing methods. Wang et al.~\cite{wang2023sstformer} propose an action recognition method that integrates RGB frames and event stream, called SSTFormer. By using a memory support Transformer network for RGB frame encoding, and a spiking neural network for raw event stream encoding, the current problems in event camera pattern recognition are solved. Wang et al.~\cite{wang2024generative} improves the extraction and compression model of spatiotemporal feature semantics by training a generative model based attention module in the feature representation stage. 

In addition to the work mentioned above, a great deal of excellent works have also been proposed in recent years. Chen et al.~\cite{chen2024ost} propose a method for optimizing spatiotemporal descriptors to improve text knowledge and promote general video recognition. It's innovation lies in enhancing action category names through a large language model. Zheng et al.~\cite{zheng2024spatio} propose a human action recognition method that achieves spatiotemporal fusion through joint trajectory maps. This method can better capture rich spatiotemporal dependencies and achieve excellent performance. Kahatapitiya et al.~\cite{kahatapitiya2024victr} propose a new video text modeling method, VicTR, which can achieve better performance in video text models by combining video and text information to generate "video conditional text" embeddings. Different from these works, in this paper, we design a novel SSM based human action recognition method to learn the visual representations effectively.

\subsection{State Space Model}  
State Space Model (SSM)~\cite{kalman1960new} is the recently popular model, which is used to achieve state space transformation. To improve the long-range modeling capability of the models, Gu et al.~\cite{gu2021combining} introduce a new deep learning model that combines the advantages of recurrent neural networks (RNNs), time convolution network, and neural differential equations (NDEs) in the form of linear state space layers (LSSL). LSSL maps sequences by simulating linear continuous time state space representations, theoretically proving its close relationship with the above three models and inheriting their advantages. Subsequently, Gu et al.~\cite{gu2021efficiently} proposes structured state space sequence models (S4) for deep learning, as a novel alternative to CNNs or Transformers. Based on the new State Space Model(SSM) parameterization, S4 can efficiently model long-range dependencies in long sequences, and it solves the computational and memory requirements encountered by existing methods when processing long sequences.

Building on these works, many researchers are beginning to explore the potential of SSM. Islam et al.~\cite{islam2022long} uses S4 as a decoder to model long-range temporal interactions in movie clips. Nguyen et al.~\cite{nguyen2022s4nd} further extends 1-dimensional sequence modeling to 2-dimensional and 3-dimensional models, such as images and videos. Subsequently, Gu et al.~\cite{gu2023mamba} introduce a selection scanning mechanism in the Mamba model, largely improving the performance of S4 on long-range sequence understanding. Furthermore, Vision Mamba~\cite{zhu2024vision} and VMamba~\cite{liu2024VMamba} successfully demonstrate the efficacy of using multiple scanning orders to enhance model performance, which uses bidirectional scanning and four-way scanning respectively. VideoMamba~\cite{li2024videomamba} innovatively applies Mamba models to the video field, overcoming the limitations of existing 3D convolutional neural networks and Transformers. 
In this paper, we design a novel scanning method along temporal trajectories for event voxels. This innovative mechanism augments the model's capacity to extract rich temporal details from event videos.

\section{Methodology}  
In this section, we will introduce how to effectively explore the task of event-based human action recognition using a Mamba-based framework. Firstly, we provide an overview to summarize the framework proposed in this work. Next, we introduce the effective representations of event data in this work. Finally, we analyze the detailed network architectures in our framework for event-based human action recognition.

\subsection{Overview}  
To explore the effectiveness of the Mamba network in event-based human action recognition tasks, in this work, we propose a novel action recognition framework based on a spatial-temporal scanning Mamba architecture. As shown in Fig.~\ref{fig:framework}, we input various forms of event data representations (event image and event voxel) into the network to fully utilize the effective information of the event data. Specifically, we propose a voxel temporal scanning mechanism to process the voxel grids by rearranging them in chronological order, which effectively integrates the dense temporal information of each event trajectory. Subsequently, we used VMamba~\cite{liu2024VMamba} as our backbone network to extract robust feature representations for both the spatial and the temporal. Finally, we add the features of the two perspectives and send them to the classification head for efficient action classification. More details will be introduced in the subsequent paragraphs.

\subsection{Input Representation}  
Owing to the rich temporal information inherent in event data, our approach involves integrating data from multi-view to harness the full potential of event data's informative representation. We consider an event stream denoted as $\mathcal{E} \in \mathbb{R}^{W \times H \times T} = \{e_1, e_2, ..., e_N\}$, each $e_i = [x, y, t, p]$ represents an individual event point triggered asynchronously, $(x, y)$ denotes the spatial coordinates, $t$ is the timestamp, $p$ is the polarity, with $i$ ranging from 1 to $N$, and $N$ signifying the total count of event points within the current sample. $H, W, T$ denotes the height, width, and overall time step. 
Utilizing the event stream $\mathcal{E}$, one can effectively compile them into event frames $\mathcal{E}_{F} \in \mathbb{R}^{T' \times C \times H \times W}$, $T'$ denotes the number of event frames, $C$ is the channel of each frame, thereby leveraging the full potential of pre-existing deep network architectures (e.g., ViT~\cite{dosovitskiy2020ViT}, Mamba~\cite{gu2023mamba}). Specifically, event images are generated by segmenting the event stream into a series of clips, each encapsulating a fixed time interval. This approach is effective for event-based recognition~\cite{wang2024hardvs, wang2024event}, however, the rich potential of the dense temporal dimension is seldom fully exploited.

In this work, we also investigate the use of the \textit{event voxel} to further enhance event representation which can effectively preserve the temporal information of the event stream. Specifically, we dissect the event stream $\mathcal{E} \in \mathbb{R}^{W \times H \times T}$ into multiple cubic voxels $\mathcal{E}_{V} \in \mathbb{R}^{a \times b \times c}$, each of which may contain several event points. Thus, we have $\frac{W}{a} \times \frac{H}{b} \times \frac{T}{c}$ voxel grids for event stream $\mathcal{E}$. In the following sub-sections, we will further explore how to obtain the voxel tokens using the temporal voxel scanning mechanism for our framework.

\subsection{Network Architecture} 

As shown in Fig.~\ref{fig:framework} (the top sub-figure), our event-based HAR framework is built based on the Mamba network which takes the event frames and voxels as the input. We will first introduce the regular Mamba network for event frame-based action recognition. Then, we will introduce the temporal voxel scan mechanism to augment the HAR framework further.

\noindent $\bullet$  \textbf{Mamba based Visual Backbone.~}  
Given the stacked event frames $\mathcal{E}_{F} \in \mathbb{R}^{T' \times C \times H \times W}$, we first partition each image into patches and re-formulate them using spatial cross-scan employed in VMamba network. Then, a patch embedding layer is adopted to transform them into token representations before feeding into the VMamba backbone network. A four-stage VSS block is adopted to construct the VMamba backbone. Specifically, Mamba~\cite{gu2023mamba} maps a 1-dimensional function or sequence $x(t) \in \mathbb{R}$ $\rightarrow$ $y(t) \in \mathbb{R}$ through a hidden state $h(t) \in \mathbb{R}^{N}$. Just like the recurrent neural network (RNN), Mamba usually inputs the previous state and the current input into the network to obtain the current output. This continuous system can be defined as,
\begin{equation}  
\begin{aligned} 
\label{continuous_SSM}
h'(t) = \mathbf{A}h(t) + \mathbf{B}x(t),    \\  
y(t) = \mathbf{C}h(t) +  \mathbf{D}x(t).
\end{aligned}  
\end{equation}
where $\mathbf{A} \in \mathbb{R}^{N \times N} $ is the state matrix, $\mathbf{B} \in \mathbb{R}^{N \times L} $ is the input matrix , $\mathbf{C} \in \mathbb{R}^{L \times N} $ is the output matrix, $\mathrm{~and~} \mathbf{D} \in \mathbb{R}^{L \times L}$ is the feed-through matrix. $x(t) \in \mathbb{R}^{L}$ and $h'(t) \in \mathbb{R}^{N}$ are the current input sequence and the derivative of the hidden state. Finally, the $y(t)$ can be obtained as the current output. For more detailed information, including details of discretization, please refer to the supplementary materials or Mamba~\cite{gu2023mamba}.



For each Mamba layer, we first input image data for feature learning from a spatial view. Fig.~\ref{fig:framework} (c) shows the detailed structure of the VMamba backbone blocks. Firstly, we take $X_i \in \mathbb{R}^{B \times H \times W \times C}$, $i \in \{image, voxel \}$ as input (voxel will be introduced in the next sub-section), where $B$ is the batch size, $H$ and $W$ denote height and width, $C$ is the dimension of feature embedding. After passing through a linear normalization (LN) layer, the input $X_i$ is divided into two parts, $x_i$ and $z_i$, according to the last dimension. Subsequently, we take $x_i$ as the input of the main branch, and after the linear layer, \textit{DW} convolution, and $\emph{SiLU}$ activation function, it is sent to the \textit{SS2D} module. The formulas can be described as,
\begin{equation}  
\begin{aligned} 
\label{VSS_Block_x}
x_i' = SS2D(\emph{SiLU}(DW(\emph{Linear}(x_i)))).  \\
\end{aligned}  
\end{equation}
The \textit{SS2D} module uses a cross-selective scanning mechanism to scan image patches with four-way repeatedly. For detailed information on four-way scanning, please refer to VMamba~\cite{liu2024VMamba}. Next, we merged the four different orders of patches to the original 2D feature map through the cross-merge operation. Then, we multiply the $x_i'$ obtained from the \textit{SS2D} module and the linear layer by $z_i$, which also through a linear layer and \textit{SiLU} activation function. Finally, we add the result of this multiplication to the initial input $X_i$ to produce the final output result of the current block. These processes can be expressed as,
\begin{equation}  
\begin{aligned} 
\label{xi_zi}
X_i' = Linear(LN(x_i') \times \emph{SiLU}(Linear(z_i)) ) + X_i.
\end{aligned}  
\end{equation}
After that, we input $X_i'$ into the block of the next layer. Through the above steps, we stacked $L$ layers of S6 blocks for effective feature extraction.

\noindent $\bullet$  \textbf{Voxel Temporal Scanning Mechanism.~} 
The aforementioned spatial cross-scan used in Mamba captures the spatial information well, but we believe the temporal information can also be further enhanced. In this work, we propose a \textit{temporal voxel scan mechanism} to augment the frame-based Mamba network for event-based human action recognition. As shown in Fig.~\ref{fig:framework} (b), given the event stream $\mathcal{E}$, we first divide it into multiple clips (i.e., blue, orange, green blocks) and extract the voxel representation for each clip. Note that, we removed voxels that hardly contain any events to save on computational costs. 
Let's take the $i$-th blue event voxel $\mathcal{E}_V^i$ as an example, we search its trajectory in the next clip (i.e., the orange one) by measuring its cosine similarity (\textit{CoSim}):
\begin{align} 
arg~~max(CoSim( \mathcal{E}_V^i , \mathcal{E}_V^{k} )), k \in \{1, 2, ..., M\},  
\end{align} 
where $M$ is the maximum number of informative event voxels in the orange clip. The orange voxel with the highest similarity will be selected, and similar operations will be conducted for other event voxels to construct the event voxel sequence $[\mathcal{E}_V^1, \mathcal{E}_V^2, ..., \mathcal{E}_V^N]$. We propose a voxel embedding layer to transform them into event tokens before feeding into the VMamba network.

\begin{figure*}[!htb]
\center
\includegraphics[width=7in]{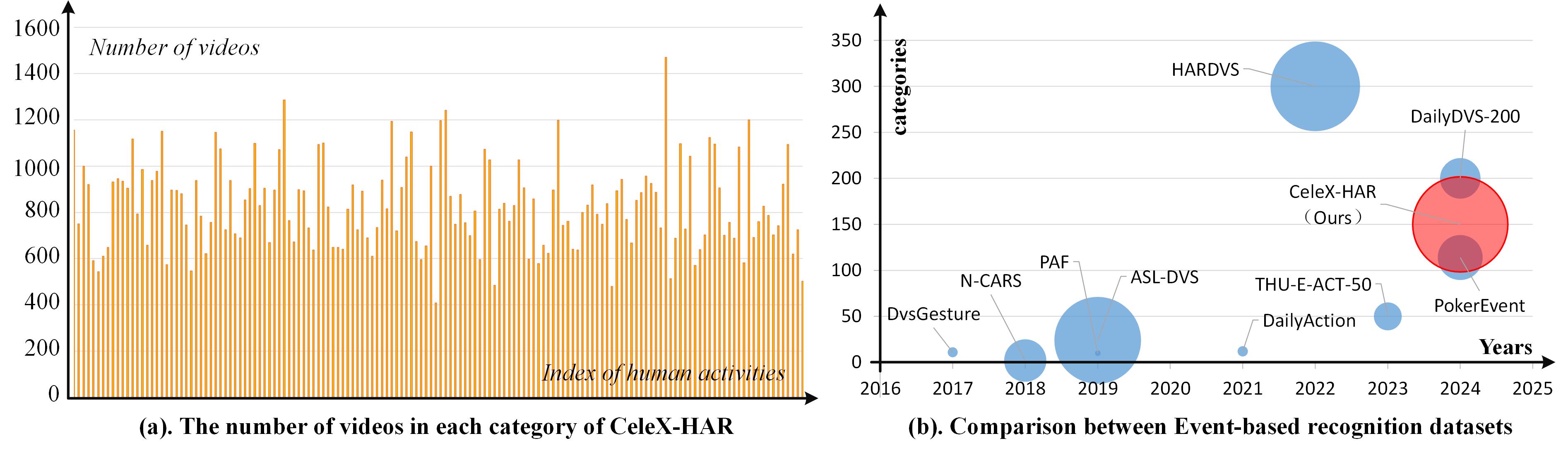} 
\caption{Information of (a). the number of videos in each category of CeleX-HAR, and (b). Comparison between existing Event-based recognition datasets.}   
\label{sampleVIS}
\end{figure*}



\noindent $\bullet$ \textbf{Classification Head.~}
After passing through the S6 blocks, we can obtain the output features from both the spatial and temporal views. 
Then, we repeat the features of the voxel into the same shape as the image features and fuse the features of the two branches through simple addition. 
Subsequently, we input the fusion features to the classification head after passing through an average pooling layer. 
Our classification head utilizes a simple linear layer to map feature dimensions and obtain predicted scores for each action category. Finally, we calculate the cross entropy loss between the predicted results and the ground truth labels, which can be formulated as: 
\begin{equation}
\label{CE_loss} 
\mathcal{L} = -\frac{1}{N}\sum_{i=1}^N y_i\log\hat{y_i} + (1-y_i)\log(1-\hat{y_i}).
\end{equation}
where $\hat{y_i}$ is the predicted scores and ${y_i}$ is the true label.

\begin{figure*}
    \centering
    \includegraphics[width=0.95\linewidth]{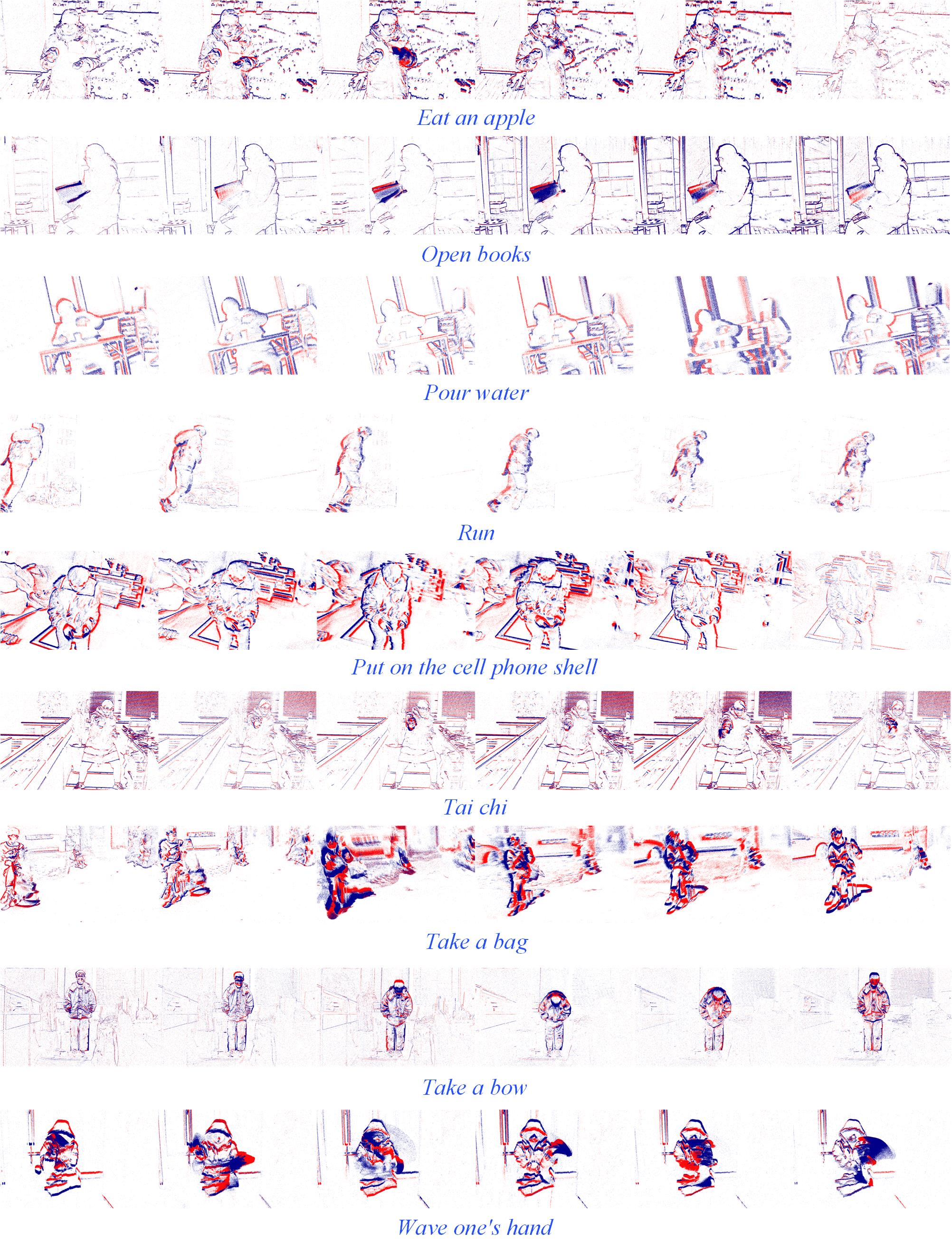}
    \caption{More illustrations of event samples in our CeleX-HAR dataset.} 
    \label{fig:CeleXHARmoreSamples}
\end{figure*}

\section{CeleX-HAR Benchmark Dataset} 

In this section, we first introduce the protocols we followed when collecting the event videos. Then, we will focus on statistical analysis and baselines we build for future works to compare.

\subsection{Protocols} 
When collecting our CeleX-HAR dataset, we obey the following standards: 
\emph{1). Multi-view}: We record each action from different views, including front, left, right, down, and up views. 
\emph{2). Various illumination}: Our dataset also contains videos with different illumination conditions, including low-, middle-, and full-illumination. Note that, the low-illumination videos account for about $40\%$ for each class.  
\emph{3). Camera motions}: The movement of the event camera affects the imaging quality greatly. As the camera moves, a great deal of background information is recorded; when the camera is still, only the moving subject is recorded. Note that half of the videos in each class are recorded using a moving event camera. 
\emph{4). Video length}: Usually, an action lasts about 2-3 seconds in our dataset; but some videos are larger than 5 if it takes longer to complete.  
\emph{5). Speed of action}: The video of the three kinds of speeds (i.e., low-, medium-, and high-speed) occupies the proportion of $30\%, 40\%, 30\%$, respectively. 
\emph{6). Occlusion}: In a certain movement, there can be partial occlusion of the interference. 
\emph{7). Glitter}: As the event camera is sensitive to glitter, in our dataset collection, we also record videos with conspicuous flash. 
\emph{8). Capture distance}: These videos are approximately 1-2 meters, 2-3 meters, and larger than 3 meters away from the target object. 
\emph{9). Large-scale \& High-definition}: 124,625 high-definition ($1280 \times 800$) event video sequences (the largest real-event HAR dataset) are recorded which cover 150 classes of human activities.

\subsection{Statistical Analysis and Baselines}  

Our dataset comprises 150 categories of common human behavioral actions, as illustrated in Fig.~\ref{sampleVIS} (a), a total of 124,625 videos are collected using a CeleX-V event camera and the training and testing subset contains 99,642 and 24,983 videos respectively. As visualized in Fig.~\ref{sampleVIS} (b), we compare the number of categories (Y-axis values) and the number of videos (bubble size) between the CeleX-HAR dataset and other datasets.

As shown in Table~\ref{CeleX_CeleXResults}, we also establish benchmarks for a variety of recognition models on our dataset, which provides a wide baseline for future comparisons on the CeleX-HAR dataset, including: 
1). \textit{CNN based models} (ResNet50~\cite{he2016deep}, ConvLSTM~\cite{shi2015convolutional}, C3D~\cite{tran2015learning}, R3Plus1D~\cite{tran2018closer}, TSM~\cite{lin2019tsm}, ACTION-Net~\cite{wang2021action}, TAM~\cite{liu2021tam}, GSF~\cite{sudhakaran2023gate}), 
2). \textit{Transformer based models} (Video-SwinTrans~\cite{liu2022video}, TimeSformer~\cite{bertasius2021space}, SlowFast~\cite{feichtenhofer2019slowfast}, SVFormer~\cite{xing2023svformer}, EFV++~\cite{chen2024retain}, ESTF~\cite{wang2024hardvs}), 
3). \textit{RWKV based models} (VRWKV~\cite{duan2024vrwkv}) and 
4). \textit{Mamba based models} (Vision Mamba~\cite{zhu2024vision}, VMamba~\cite{liu2024VMamba}, Video Mamba~\cite{li2024videomamba}).

\section{Experiments} 

\subsection{Dataset and Evaluation Metric}  

In this paper, our experiments are conducted on the \textbf{ASL-DVS}~\cite{bi2020graph}, \textbf{N-Caltech101}~\cite{orchard2015converting}, \textbf{DVS128-Gait-Day}~\cite{wang2021eventgait3Dgraph}, \textbf{Bully10K}~\cite{dong2024bullying10k}, \textbf{Dailydvs-200}~\cite{wang2024dailydvs}, and our newly proposed \textbf{CeleX-HAR} dataset. More detailed introductions to these datasets can be found in our Supplementary Materials. 
The \textbf{top-1} accuracy is adopted for the evaluation of our proposed model and other SOTA action recognition algorithms.

\subsection{Implementation Details} 
Our proposed framework can be optimized in an end-to-end manner. The learning rate and weight decay are set as 0.001 and 0.0001, respectively. The SGD is selected as the optimizer and trained for a total of 30 epochs. In our implementations, a total of $33$ S6 blocks are stacked as our backbone network like VMamba-B~\cite{liu2024VMamba}. We rich the temporal information of the input through the implementation of a novel temporal scanning strategy. Besides, we select 8 event frames as the input of the event images, as other benchmarked baselines.  
Our code is implemented using Python based on PyTorch~\cite{paszke2019pytorch} framework and the experiments are conducted on a server with CPU Intel(R) Xeon(R) Gold 5318Y CPU @2.10GHz and GPU RTX3090s. More details can be found in our source code.

\subsection{Comparison with Other SOTA Algorithms}

\noindent $\bullet$ \textbf{Results on ASL-DVS~\cite{bi2020graph} Dataset.~} 
As shown in Table~\ref{ASLDVSResults}, our proposed method achieves 99.9\% (top-1 accuracy) on the ASL-DVS dataset. The compared method M-LSTM which adopts learnable event representation is still inferior to our method. Some graph-based event recognition models are also worse than ours, including EV-VGCNN, VMV-GCN, and Ev-Gait-3DGraph. Therefore, we can conclude that our proposed model is more effective for event-based HAR.


\begin{table}[!htp]
\center
\scriptsize   
\caption{Results on the ASL-DVS~\cite{bi2020graph} Dataset.~} 
\label{ASLDVSResults}
\resizebox{0.48\textwidth}{!}{
\begin{tabular}{ccccc} 		
\hline 
\textbf{EST} &\textbf{AMAE}     &\textbf{M-LSTM}    &\textbf{MVF-Net}    & \textbf{EventNet}  \\  
0.979   & 0.984     &0.980     &0.971     &0.833    \\ 
\hline 
\textbf{RG-CNNs}   &\textbf{EV-VGCNN}   &\textbf{VMV-GCN}   &\textbf{EV-Gait-3DGraph}   &\textbf{Ours} \\
0.901     &0.983     &0.989  &0.738  &\textbf{0.999}	 \\
\hline
\end{tabular} } 
\end{table}


\begin{table}
\center
\scriptsize  
\caption{Results on N-Caltech101~\cite{orchard2015converting} Dataset.~} 
\label{Caltech101Results} 
\setlength\tabcolsep{3.3pt}
\resizebox{0.48\textwidth}{!}{
\begin{tabular}{cccccccc} 		
\hline
\textbf{EventNet}    &\textbf{RG-CNNs}     &\textbf{VMV-GCN}     &\textbf{EV-VGCNN}     &\textbf{EST}     &\textbf{ResNet-50}  \\
0.425       &0.657     &0.778     &0.748       &0.753     &0.637   		 \\
\hline
\textbf{MVF-Net}     &\textbf{M-LSTM}      &\textbf{AMAE}        &\textbf{ESTF}         &\textbf{EFV++}   &\textbf{Ours}     	  		 \\ 
0.687      &0.738      &0.694     &0.832       &0.897     &\textbf{0.939}    	  		 \\
\hline
\end{tabular}
} 
\end{table}

\noindent $\bullet$ \textbf{Results on N-Caltech101~\cite{orchard2015converting} Dataset.~} 
As shown in Table~\ref{Caltech101Results}, our model achieves 93.9\% (top-1 accuracy) on this benchmark dataset which is significantly better than the compared methods. To be specific, our model outperforms ResNet50 by +30\% on the top-1 accuracy metric and surpasses the latest state-of-the-art method EFV++ by +4.2\%. Due to the powerful modeling ability of our model and the important role of temporal information, our model achieves superior performance which also validates its effectiveness. 

\begin{table}
\center
\scriptsize   
\caption{Results on the DVS128-Gait-Day~\cite{wang2021eventgait3Dgraph} dataset.~} 
\label{DVS128GaitResults} 
\resizebox{0.48\textwidth}{!}{
\begin{tabular}{ccccccc} 		
\hline 
\textbf{EVGait-3DGraph}   &\textbf{2DGraph-3DCNN} &\textbf{EV-Gait-IMG} \\ 
94.9   &92.2     &87.3 \\
\textbf{LSTM-CNN}   &\textbf{SVM-PCA}  &\textbf{Ours}  \\
86.5     &78.1     &\textbf{99.0}    	  		 \\
\hline 
\end{tabular}
} 
\end{table} 


\noindent $\bullet$ \textbf{Results on DVS128-Gait-Day~\cite{wang2021eventgait3Dgraph} Dataset.~} 
This dataset is specifically proposed for human gait recognition by Wang et al., as shown in Table~\ref{DVS128GaitResults}, we can find that the EVGait-3DGraph already achieves $94.9\%$ (acc/top-1) on this dataset. In contrast, our proposed method obtains 99.0\% on acc/top-1 which is much better than the 3DGraph based recognition model. The outstanding results on this dataset fully demonstrate that our model works well on event-based recognition.

\noindent $\bullet$ \textbf{Results on Bully10K~\cite{dong2024bullying10k} Dataset.~}
As shown in Table~\ref{Bullying10k_acc}, we report the experimental results on the Bully10K dataset. It can be seen that our method has reached the SOTA level among many methods (such as HRNet~\cite{sun2019deep}, etc.). Specifically, our model also beats the latest state-of-the-art method EFV++~\cite{chen2024retainblendexchangequalityaware} which ranks in second place. However, if we lose the voxel temporal information, the results will be decreased. These results demonstrate that our method can enhance the event-based recognition results by learning from multi-view input.

\begin{table}[!htp]
\center
\caption{Results on the Bully10K~\cite{dong2024bullying10k} dataset.~} 
\label{Bullying10k_acc}
\resizebox{0.48\textwidth}{!}{ 
\begin{tabular}{l|c|c|c}
\hline \toprule [0.5 pt]
\textbf{Algorithm} & \textbf{Source}      &\textbf{Backbone}  &\textbf{acc/top-1}  \\
\hline
C3D~\cite{ji20123d} & ICCV-2015     &3D CNN   &71.25     \\
R2Plus1D~\cite{tran2018closer} & CVPR-2018   &ResNet18   &69.25  \\
R3D~\cite{tran2017convnet} &arXiv-2017   &ResNet18   &72.50    \\
TAM~\cite{liu2021tam} &  ICCV-2021    &ResNet50    &71.20     \\
SlowFas~t\cite{feichtenhofer2019slowfast} & ICCV-2019      &ResNet50   &74.00  \\ 
SNN~\cite{fang2021deep}  &   NeurIPS-2021    &SEW-ResNet19  &67.05   \\ 
X3D~\cite{feichtenhofer2020x3d}  &  CVPR-2020       &ResNet   &76.90    \\
SimpleBaseline~\cite{xiao2018simple}  &    ECCV-2018     &ResNet50   &88.30     \\
HRNet~\cite{sun2019deep}  &   CVPR-2019      &HRNet-ws32   &88.20     \\
EFV++~\cite{chen2024retainblendexchangequalityaware}     &arXiv-2024  &Former-GNN   &90.51    \\ 
\hline
\textbf{Ours}       &-     &SSM   &\textbf{91.52}    \\
\textbf{Ours $w/o$ Voxel Scan}       &-     &SSM   &\textbf{91.17}    \\
\hline \toprule [0.5 pt]  
\end{tabular}}
\end{table}

\noindent $\bullet$ \textbf{Results on Dailydvs-200~\cite{wang2024dailydvs} Dataset.~}
The Dailydvs-200 is also a newly proposed dataset that contains 200 categories. As shown in Table~\ref{DailyDvs_result}, our EVMamba is also superior to all methods on the Dailydvs-200 dataset. Specifically, our approach achieves a top-1 accuracy improvement of +8.78\% over the classical CNN-based model TSM. Furthermore, compared to Swin-T~\cite{liu2022video}, which previously held the top rank, our model has significantly surpassed it and achieved a new SOTA level. Also, when we abandoned the voxel scan, the experimental accuracy decreased by 0.42, which further highlights the importance of voxel temporal information.

\begin{table} 
\center 
\small      
\caption{Results on the Dailydvs-200~\cite{wang2024dailydvs} Dataset.}
\label{DailyDvs_result}
\begin{tabular}{l|c|c|ccccc} 
\hline \toprule [0.5 pt]
\textbf{Algorithm}  &\textbf{Backbone}   &\textbf{acc/top-1}   &\textbf{acc/top-5}  \\
\hline 
C3D~\cite{tran2015learning}                  &3D CNN    &21.99  &45.81  \\
\hline 
I3D~\cite{ji20123d}                  &ResNet50    &32.30  &59.05   \\
\hline 	
R2Plus1D~\cite{tran2018closer}             &ResNet34     &36.06  &63.67    \\
\hline 	
SlowFast~\cite{feichtenhofer2019slowfast}             &ResNet50     &41.49  &68.19      \\
\hline 
TSM~\cite{lin2019tsm}                  &ResNet50     &40.87  &71.46   \\
\hline 	
EST~\cite{gehrig2019end}           &ResNet34     &32.23  &59.66     \\
\hline 
TimeSformer~\cite{bertasius2021space}               &Transformer    &44.25  &74.03     \\
\hline 	
Swin-T~\cite{liu2022video}                     &Transformer     &48.06  &74.47          \\
\hline 
ESTF~\cite{wang2024hardvs}                 &ResNet18     &24.68  &50.18    \\
\hline 
GET~\cite{peng2023get}                 &Transformer     &37.28  &61.59    \\
\hline 
Spikingformer~\cite{zhou2022spikformer}                 &Transformer    &36.94  &62.37    \\
\hline 
SDT~\cite{yao2024spike}                 &Transformer    &35.43  &58.81    \\
\hline 
\textbf{Ours}                 &SSM     &\textbf{49.65}     &\textbf{75.89}       \\
\hline
\textbf{Ours $w/o$ Voxel Scan}    &SSM     &\textbf{49.23}     &\textbf{75.67}       \\
\hline \toprule [0.5 pt] 
\end{tabular}  
\end{table}

\begin{table*}[!htp]
\small      
\center 
\caption{Experimental results on CeleX-HAR dataset.~}  
\label{CeleX_CeleXResults}
\begin{tabular}{c|l|c|c|c|c|c|c} 
\hline \toprule [0.5 pt] 
\textbf{No.} &\textbf{Algorithm}  &\textbf{Publish}  &\textbf{Arch.} &\textbf{FLOPs}  &\textbf{Params}  &\textbf{acc/top-1}  &\textbf{Code} \\ 
\hline 
01 & ResNet-50~\cite{he2016deep}  &CVPR-2016    &CNN  &8.6G  &11.7M  &0.642   &\href{https://github.com/KaimingHe/deep-residual-networks}{URL}      \\	
02 & ConvLSTM~\cite{shi2015convolutional}  &NIPS-2015     &CNN, LSTM   &-  &- &0.539   &\href{https://github.com/ndrplz/ConvLSTM_pytorch}{URL}      \\		
03 & C3D~\cite{tran2015learning}      &ICCV-2015      &CNN  &0.1G &147.2M &0.630  &\href{https://github.com/leftthomas/R2Plus1D-C3D}{URL}       \\	
04 & R2Plus1D~\cite{tran2018closer}  &CVPR-2018     &CNN  &20.3G &63.5M &0.679    &\href{https://github.com/leftthomas/R2Plus1D-C3D}{URL}       \\
05 & TSM~\cite{lin2019tsm}      &ICCV-2019     &CNN  &0.3G &24.3M  &0.704  &\href{https://github.com/mit-han-lab/temporal-shift-module}{URL}      \\	
06 & ACTION-Net~\cite{wang2021action}  &CVPR-2021      &CNN &17.3G &27.9M  &0.685   &\href{https://github.com/V-Sense/ACTION-Net}{URL}      \\
07 & TAM~\cite{liu2021tam} 	  &ICCV-2021        &CNN  &16.6G &25.6M  &0.705   &\href{https://github.com/liu-zhy/temporal-adaptive-module}{URL}       \\
08 & GSF~\cite{sudhakaran2023gate} &TPAMI-2023  &CNN   &16.5G     &10.5M       & 0.703         &\href{https://github.com/swathikirans/GSF}{URL}  \\
\hline 
09 & V-SwinTrans~\cite{liu2022video} &CVPR-2022      &ViT  &8.7G &27.8M  &0.689  &\href{https://github.com/SwinTransformer/Video-Swin-Transformer}{URL}       \\
10 & TimeSformer~\cite{bertasius2021space}   &ICML-2021       &ViT   &53.6G &121.2M &0.680  &\href{https://github.com/facebookresearch/TimeSformer}{URL}       \\
11 & SlowFast~\cite{feichtenhofer2019slowfast} &ICCV-2019      &ViT   &0.3G &33.6M &0.680   &\href{https://github.com/facebookresearch/SlowFast}{URL}      \\
12 & SVFormer~\cite{xing2023svformer}    &CVPR-2023     &ViT   &196.0G   &121.3M   &0.610    &\href{https://github.com/ChenHsing/SVFormer}{URL}  \\
13 & EFV++~\cite{chen2024retain}   &arXiv-2024      &ViT, GNN   &36.3G   &39.2M    &0.695  &\href{https://github.com/Event-AHU/EFV_event_classification/tree/EFVpp}{URL}      \\
14 & ESTF~\cite{wang2024hardvs}  &AAAI-2024  &ViT, CNN   & 17.6G       & 46.7M   &0.673       &\href{https://github.com/Event-AHU/HARDVS}{URL}  \\
\hline 
15 & VRWKV-S~\cite{duan2024vrwkv}  &arXiv-2024  &RWKV  &4.6G   &23.8M    &0.661   &\href{https://github.com/OpenGVLab/Vision-RWKV}{URL}  \\
16 & VRWKV-B~\cite{duan2024vrwkv}  &arXiv-2024  &RWKV  &18.2G   &93.7M   &0.668  &\href{https://github.com/OpenGVLab/Vision-RWKV}{URL}   \\
\hline
17    &Vision Mamba-S~\cite{zhu2024vision}  &ICML-2024       &SSM   &5.1G   &26.0M    &0.701&\href{https://github.com/hustvl/Vim}{URL}     \\
18    &VMamba-S~\cite{liu2024VMamba}  &arXiv-2024       &SSM   &11.2G   &44.7M    &0.713   &\href{https://github.com/MzeroMiko/VMamba}{URL}     \\
19    &VMamba-S(V2)~\cite{liu2024VMamba}  &arXiv-2024       &SSM   &8.7G   &50.4M    &0.715   &\href{https://github.com/MzeroMiko/VMamba}{URL}      \\
20    &VMamba-B~\cite{liu2024VMamba}   &arXiv-2024      &SSM   &18.0G   &76.5M    &0.720 &\href{https://github.com/MzeroMiko/VMamba}{URL}       \\
21    &VMamba-B(V2)~\cite{liu2024VMamba}  &arXiv-2024       &SSM   &15.4G   &88.9M    &0.718  &\href{https://github.com/MzeroMiko/VMamba}{URL}      \\
22    &VideoMamba-S~\cite{li2024videomamba}   &ECCV-2024   &SSM  &4.3G  &26.0M    &0.669 &\href{https://github.com/OpenGVLab/VideoMamba}{URL}      \\
23   &VideoMamba-M~\cite{li2024videomamba}   &ECCV-2024   &SSM  &12.7G  &74.0M  &0.691 &\href{https://github.com/OpenGVLab/VideoMamba}{URL}      \\
\hline
24   &\textbf{EVMamba (Ours)}   &-      &SSM   &37.2G             &76.5M    &\textbf{0.723}     &-     \\
25   &\textbf{EVMamba $w/o$ Voxel Scan}   &-      &SSM   &18.0G   &76.5M    &0.720   &-     \\
\hline \toprule [0.5 pt]  
\end{tabular}
\end{table*}

\noindent $\bullet$ \textbf{Results on CeleX-HAR Dataset.~}  
Note that, our proposed model (acc/top-1: 72.3\%) has reached the SOTA level among numerous methods. Compared with TSM, our method has improved +1.9 on acc/top-1. Meanwhile, for some Transformer-based methods (e.g., TimeSformer) our methods are also superior to them. VRWKV is a recently proposed visual model, and our method also has a higher performance on top-1 accuracy compared to VRWKV-B. Furthermore, compared to some Mamba-based models that have emerged recently, it is evident that our method's top-1 accuracy is approximately two points higher than that of Vision Mamba, and it also outperforms VMamba-B. The above experiments have demonstrated the superior performance of our method to some extent, which indicates that our EVMamba model is an effective event-based human action recognition method.

\subsection{Component Analysis} 
To verify the effectiveness of our proposed method, as shown in Table~\ref{CeleX_CeleXResults}, we compared our algorithm EVMamba with the version without incorporating voxel temporal scanning. To further enhance the robustness of the Mamba model, we study how to bring dense temporal information to the input based on the image spatial scanning. Ultimately, our results were further improved with the voxel temporal scanning mechanism, achieving a top-1 accuracy of 72.3\%. Thus, we believe that the Mamba model is a good backbone replacement solution for event-based human action recognition. Meanwhile, our voxel temporal scanning strategy can further improve the accuracy of the model.

\subsection{Ablation Study}

\begin{table}[h]
\centering
\small
\caption{Ablation Studies of Event Representation, Fusion Methods, Threshold of Trajectory's Length, Number of Input Frames and Input Resolution on CeleX-HAR dataset.}
\label{ablation_study}
\resizebox{0.4\textwidth}{!}{ 
\begin{tabular}{l|c}
\hline \toprule [0.5 pt] 
\textbf{\# Event Representation} & \textbf{acc/top-1}  \\
\hline
1. Event Frame              & 0.720  \\
2. Event voxel              & 0.451  \\
3. Both Frame and voxel     & 0.723  \\
\hline
\textbf{\# Fusion Methods} & \textbf{acc/top-1} \\
\hline
1. Add & 0.723 \\
2. Concatenate & 0.717   \\
3. \textbf{A}-matrix interaction & 0.716 \\
4. \textbf{B}-matrix interaction & 0.717  \\
\hline
\textbf{\# Threshold of Trajectory's Length} & \textbf{acc/top-1} \\ 
\hline
1. $\delta$ = 1                 & 0.723  \\
2. $\delta$ = 3                 & 0.715   \\
3. $\delta$ = 9                 & 0.712  \\
\hline 
\textbf{\# Number of Input Frames} & \textbf{acc/top-1}  \\
\hline
1. $4$ frames  & 0.698   \\
2. $8$ frames  & 0.723    \\
3. $12$ frames & 0.708   \\
\hline
\textbf{\# Input resolution} & \textbf{acc/top-1} \\
\hline
1. $192 \times 192$   & 0.704   \\
2. $224 \times 224$   & 0.723   \\
3. $256 \times 256$   & 0.720     \\
4. $320 \times 200$   & 0.716   \\
\hline \toprule [0.5 pt] 
\end{tabular}
}
\end{table}


\noindent $\bullet$ \textbf{Analysis of Different Event Representations for HAR.~} 
In our experiments, we investigate different event representations for the human action recognition task. 
As shown in Table~\ref{ablation_study}, we can see that considering only the event voxel (acc/top-1, 0.451) as input would result in a significant discrepancy in the outcome compared to the event frames (acc/top-1, 0.720). Despite event voxel's commendable spatio-temporal information integration, their omission of the intrinsic polarity of events hampers a comprehensive feature portrayal. In contrast, we opted to incorporate the event voxels' temporal information into the event frames and achieve outstanding performance.

\begin{figure}
    \centering
    \includegraphics[width=1\linewidth]{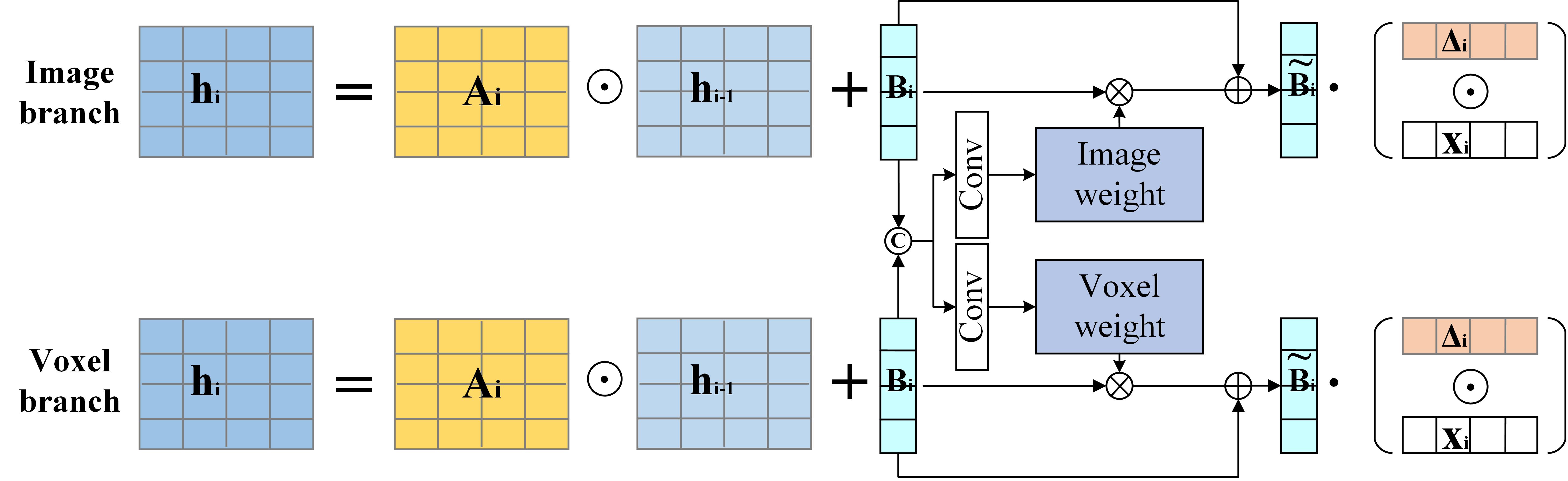}
    \caption{The fusion method of matrix interaction.}
    \label{fig:matrix_interaction}
\end{figure}

\noindent $\bullet$ \textbf{Analysis of Other Exploited Fusion Methods.~} 
Since our model's input is in the form of dual branch data (event image and voxel), we need to design a suitable scheme to integrate dense voxel temporal information into the images. Specifically, through careful study of the Mamba model, we propose a matrix interaction method as shown in Fig.~\ref{fig:matrix_interaction}. We interact with the input matrix of the image branch and the voxel branch, namely the $B$ matrix, in a gate-controlled manner to learn complementary information between the two branches. However, as shown in Table~\ref{ablation_study}, it is disappointing that this fusion method will result in a decrease in accuracy. The potential reason for this may be that the matrix interaction method breaks the original information and brings noise to the output state. Surprisingly, the simplest addition can bring the best results.

\noindent $\bullet$ \textbf{Analysis of the Threshold $\delta$ of Minimal Length of Event Temporal Voxels.~}  
When conducting the temporal event voxel scanning, we set a threshold $\delta$ to check the influence of trajectory length for the final recognition performance. In other words, we filter out the trajectories whose length is less than the threshold $\delta$. 
Specifically, the threshold $\delta$ is set as 1, 3, and 9, as shown in Table~\ref{ablation_study}, we can find that the best results can be achieved when threshold $\delta$ is 1. That is to say, it will be better when all our temporal event voxels are used. This experiment fully demonstrate the effectiveness of event voxels obtained using our temporal scan mechanism.

\noindent $\bullet$ \textbf{Analysis of Number of Input Frames.~}  
To check the influence of the input event frames, in this experiment, we input $4$, $8$, and $12$ event frames into the VMamba backbone network, and the corresponding results are 69.8\%, 72.3\%, and 70.8\% respectively. A better result can be obtained when eight frames are used. Fewer video frames can result in the model being unable to learn more effective video information, leading to underfitting. An excessive number of frames might lead to model complexity, making training more difficult and potentially increasing the risk of overfitting. This phenomenon is also referred to as information overload.

\begin{figure*}[!htp]
\center
\includegraphics[width=7in]{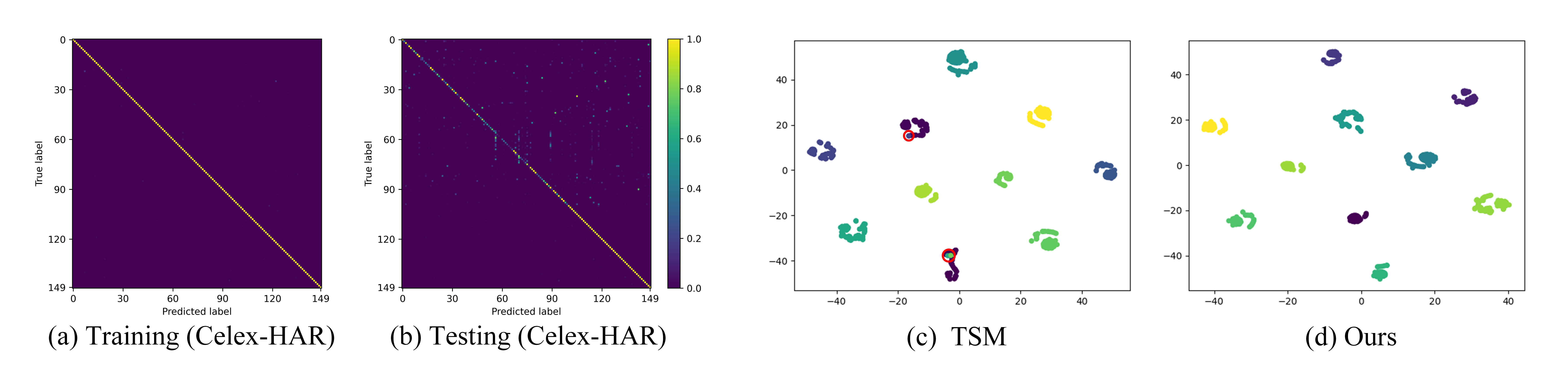} 
\caption{Visualization of the confusion matrix and feature distribution on the CeleX-HAR dataset.}   
\label{scatter_plot&confusion_matrix}
\end{figure*} 

\begin{figure}
\center
\includegraphics[width=3.3in]{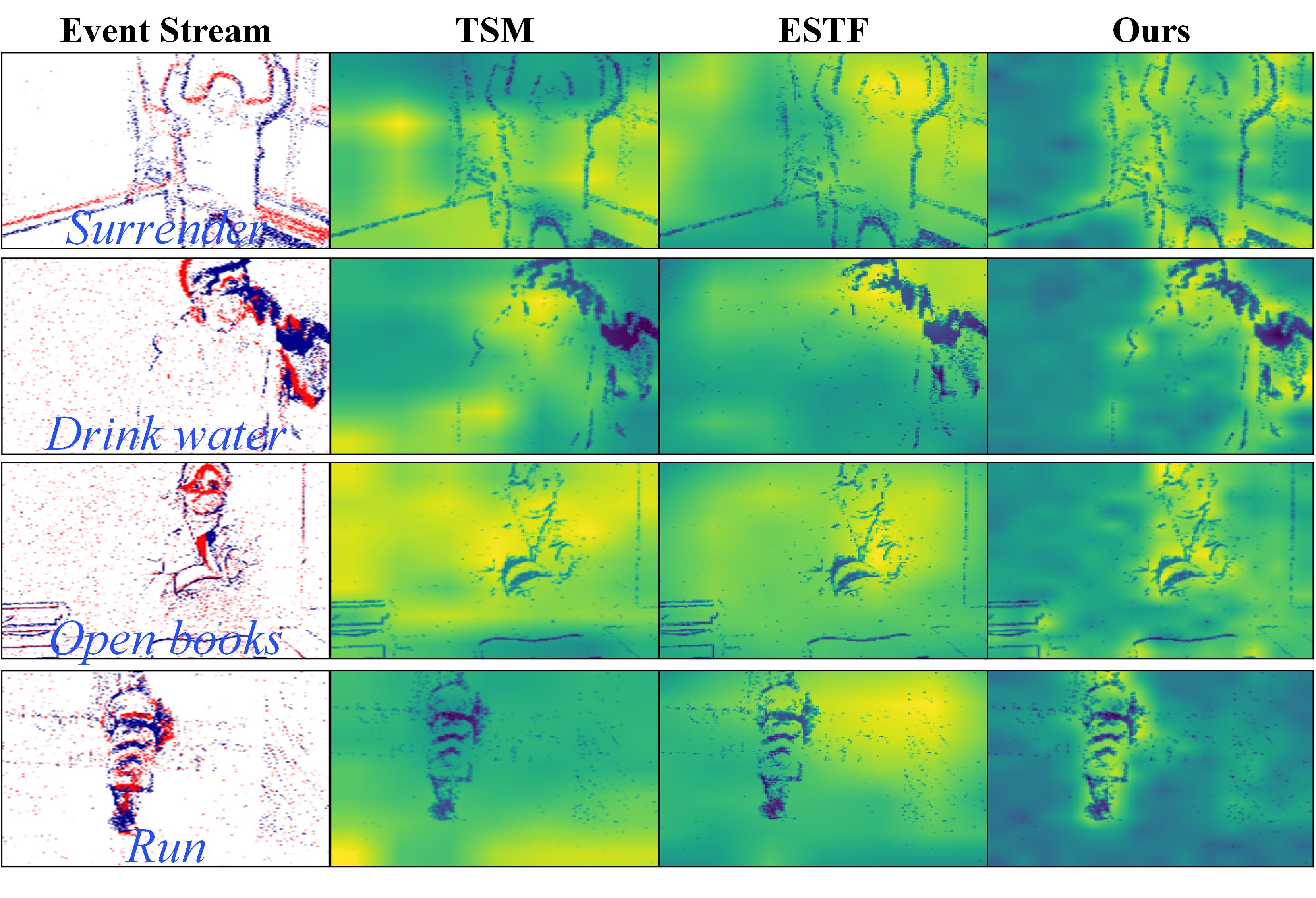} 
\caption{The feature maps obtained from TSM, ESTF, and our EVMamba.}   
\label{feature_map}
\end{figure}

\begin{figure*}
\center
\includegraphics[width=6.8in]{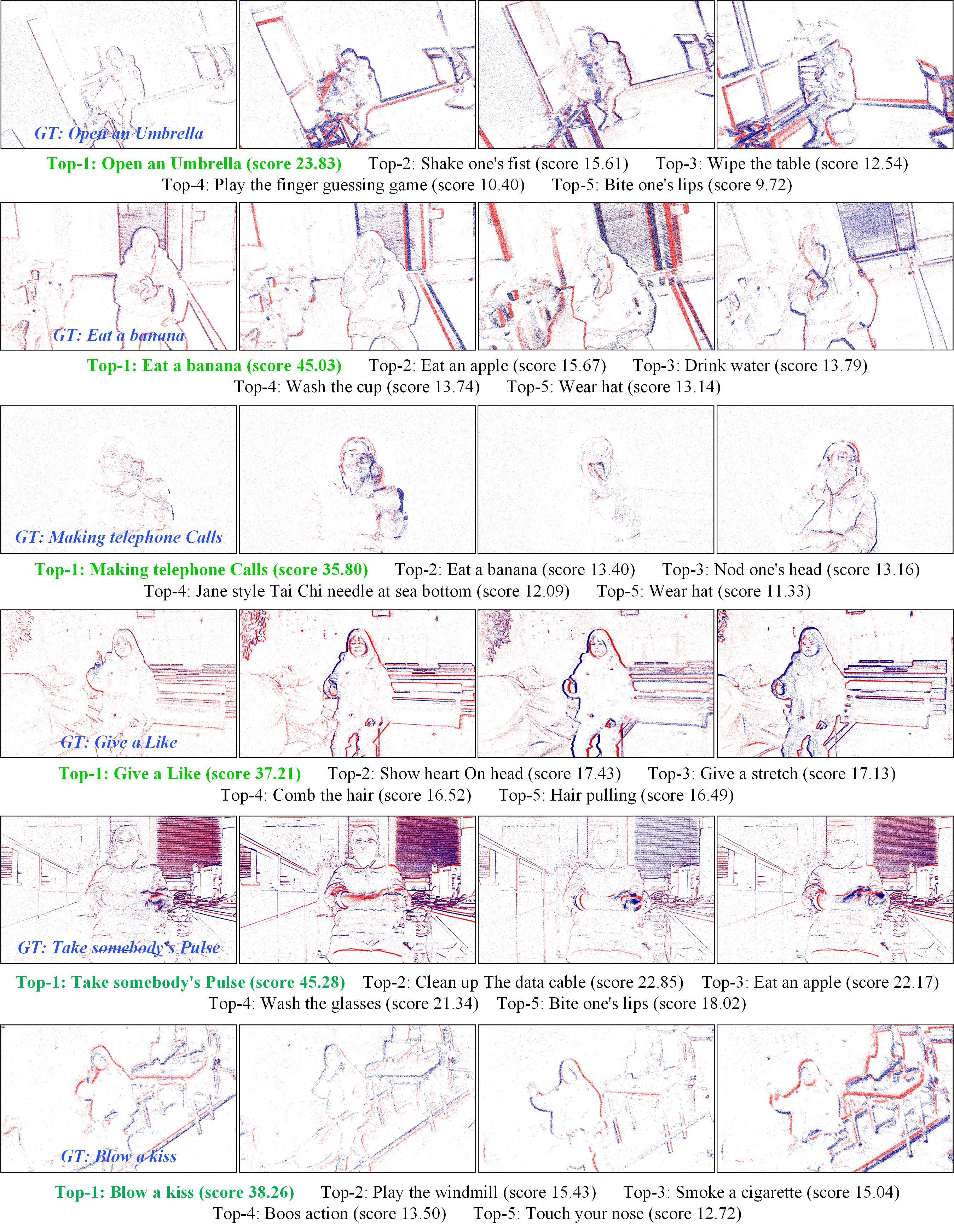} 
\caption{Top-5 predictions of our EVMamba model on CeleX-HAR dataset.}   
\label{class_top5}
\end{figure*} 

\noindent $\bullet$ \textbf{Analysis of Input Resolution of Event Stream.~}  
As this paper mainly proposes a high-definition dataset for event stream based HAR, we attempt to input frames of different resolutions, i.e., $192 \times 192$, $224 \times 224$, $256 \times 256$, and $320 \times 200$ respectively. From Table~\ref{ablation_study}, it can be observed that the highest accuracy is achieved when the input resolution is $224 \times 224$. Even though the increase in resolution introduces more tokens, it leads to a decrease in accuracy. From our experiment and other researches related to high-resolution based recognition~\cite{liu2024VMamba}, it can be observed that although high-resolution data contains a richer amount of information, the current mainstream models do not perform well. This is a very interesting and thought-provoking research problem. We speculate that when deep learning first became popular, considering the computational power at the time, successful models were experimented on with a resolution of $224 \times 224$, and these model parameters were configured for low-resolution data. With the advancement of computing resources and sensors, high-definition data is becoming more and more common. If we continue to use low-resolution data and discard the rich information of high-definition data, it is clearly a waste of resources. We hope that our newly proposed high-definition dataset can inspire and support research in this area.

\subsection{Visualization}

\noindent $\bullet$ \textbf{Confusion Marix on CeleX-HAR. }
As shown in Fig.~\ref{scatter_plot&confusion_matrix} (a) and (b), we present the confusion matrices for both training and testing on the CeleX-HAR dataset. The diagonal elements, which are closer to yellow, indicate more accurate predictions. It is evident that the model performs better on the training set compared to the testing set, highlighting the challenges posed by our dataset and suggesting that further improvements are needed for our model's generalization.

\noindent $\bullet$ \textbf{Feature Distribution. }
As depicted in Fig.~\ref{scatter_plot&confusion_matrix} (c) and (d), we also visualize the feature distributions of  TSM~\cite{lin2019tsm} and the method proposed in this paper. A total of 10 classes are randomly selected to illustrate the feature distribution of the CeleX-HAR dataset. It can be seen that our method outperforms TSM in cases of performance errors, successfully aggregating the features of each class.

\noindent $\bullet$ \textbf{Feature Map. }
As presented in Fig.~\ref{feature_map}, we concurrently display the feature maps derived from TSM, ESTF, and our model, and add them to the raw images. Specifically, we process the output features by channelizing them, resizing, and normalizing them to align with the dimensions of the raw images, and subsequently superimpose them onto the raw images. As can be observed from the figures, the feature maps generated by our model exhibit a heightened focus on the target.

\noindent $\bullet$ \textbf{Top-5 Recognition Results. } 
As shown in Fig.~\ref{class_top5}, the top-5 prediction results on our proposed CeleX-HAR dataset are presented. We present six action categories including \emph{Open an Umbrella, Eat a banana, Making telephone calls, Give a Like, Take somebody's Pulse} and \emph{Blow a kiss}. It is clear that the highest scores predicted by our model for these actions accurately correspond to their respective labels.

\subsection{Limitation Analysis}  
Although our model can achieve better accuracy compared to dense Transformer models and has advantages in efficiency, processing high-resolution event data directly still poses computational cost issues. The training and inference of the model still require high-end GPUs to complete. Therefore, in future work, we will consider techniques such as model distillation and quantization to further reduce the inference cost. Due to the limitation of GPU memory, we are unable to complete the model training and testing on the original high-definition ($1280 \times 800$) event stream data at this stage. In future work, we will design new deep neural network models for efficient high-definition data perception.

\section{Conclusion}  
This paper has made significant contributions to the field of Human Action Recognition (HAR) using event cameras. We have introduced CeleX-HAR, a large-scale, high-definition event-based HAR dataset that addresses the limitations of existing low-resolution datasets. With 150 action categories and 124,625 video sequences, CeleX-HAR considers various challenging factors such as multi-view, illumination, action speed, and occlusion, providing a comprehensive benchmark for future research. Additionally, we have proposed EVMamba, a novel event stream-based HAR model that leverages a Mamba vision backbone with innovative spatio-temporal scanning mechanisms. EVMamba has demonstrated promising performance across multiple datasets, further advancing the state-of-the-art in event-based HAR. With the release of both the dataset and source code upon acceptance, we anticipate that this work will serve as a valuable resource and foundation for future developments in the field of event-based HAR and computer vision. In our future works, we will focus on further reducing the model complexity to handle higher-resolution event data.

{
    \small
    \bibliographystyle{ieeenat_fullname}
    \bibliography{reference}
}

\end{document}